\documentclass[11pt,a4paper]{article}
\usepackage[hyperref]{acl2021}
\usepackage{times}
\usepackage{latexsym}

\usepackage{longtable}
\usepackage{multirow}

\usepackage{graphicx}
\usepackage{multirow, multicol}
\usepackage[colorinlistoftodos,prependcaption]{todonotes}
\usepackage{tabulary,booktabs}
\usepackage{todonotes}
\usepackage{subfig}
\usepackage{tabulary,booktabs}
\usepackage{tablefootnote}
\usepackage{amsfonts}
\usepackage{color,soul}
\usepackage{multirow}
\usepackage{amsmath}
\usepackage{colortbl}
\usepackage{color, xcolor}
\usepackage{bm}
\usepackage[subtle]{savetrees}
\usepackage{comment}
\usepackage{amsmath}
\usepackage{color,soul}
\usepackage{adjustbox}
\usepackage{pbox}
\usepackage{float}
\usepackage{pifont}
\usepackage{chngpage}

\usepackage{xspace}

\usepackage{microtype}

\aclfinalcopy 

\newcommand{\afterinternship}[1]{}
\newcommand{\method}{\textsc{HyperFormer}\xspace}
\newcommand{\methodefficient}{\textsc{HyperFormer}\texttt{++}\xspace}
\newcommand{\methodefficientbase}{\textsc{HyperFormer}\texttt{++}\textsubscript{\tiny BASE}\xspace}
\newcommand{\methodefficientsmall}{\textsc{HyperFormer}\texttt{++}\textsubscript{\tiny SMALL}\xspace}
\newcommand{\methodbase}{\textsc{HyperFormer}\textsubscript{\tiny BASE}\xspace}
\newcommand{\methodsmall}{\textsc{HyperFormer}\textsubscript{\tiny SMALL}\xspace}

\newcommand{\madapter}{Adapters$\dagger$\xspace}
\newcommand{\madaptersmall}{Adapters$\dagger$\textsubscript{\tiny SMALL}\xspace}
\newcommand{\madapterbase}{Adapters$\dagger$\textsubscript{\tiny BASE}\xspace}
\newcommand{\adapter}{Adapters\xspace}
\newcommand{\adaptersmall}{Adapters\textsubscript{\tiny SMALL}\xspace}
\newcommand{\adapterbase}{Adapters\textsubscript{\tiny BASE}\xspace}
\newcommand{\basesmall}{T5\textsubscript{\tiny SMALL}\xspace}
\newcommand{\basebase}{T5\textsubscript{\tiny BASE}\xspace}
\newcommand{\madapteritalic}{\textit{Adapters}$\dagger$\xspace}

\newcommand{\glue}{\textsc{GLUE}\xspace}

\newcommand{\todocomment}[1]{} 

\newcommand{\size}{\scriptsize}
\newcommand{\std}[1]{{\size$\pm$#1}}

\newcommand{\eqend}{\\[-4.4ex]\nonumber} 
\newcommand{\myspace}{\vspace{-0.8ex}}

\title{Parameter-efficient Multi-task Fine-tuning for Transformers\\ via Shared Hypernetworks}

  \author{Rabeeh Karimi Mahabadi\thanks{~~Work done while the author was at Google.} \\
  EPFL University, Idiap Research Institute\\
  \texttt{rabeeh.karimimahabadi@epfl.ch} \\\And
  Sebastian Ruder \\
  DeepMind \\
  \texttt{ruder@google.com} \\\AND
  Mostafa Dehghani \\
  Google Brain \\
  \texttt{dehghani@google.com} \\\And
  James Henderson \\
  Idiap Research Institute \\
  \texttt{james.henderson@idiap.ch}\\
  }

\date{}

\begin{document}
\maketitle

\begin{abstract} 
State-of-the-art parameter-efficient fine-tuning methods rely on introducing adapter modules between the layers of a pretrained language model. However, such modules are trained separately for each task and thus do not enable sharing information across tasks.
In this paper, we show that we can learn adapter parameters for all layers and tasks by generating them using shared hypernetworks, which condition on task, adapter position, and layer id in a transformer model. This parameter-efficient multi-task learning framework allows us to achieve the best of both worlds by sharing knowledge across tasks via hypernetworks while enabling the model to adapt to each individual task through task-specific adapters.
Experiments on the well-known GLUE benchmark show improved performance in multi-task learning while adding only $0.29\%$ parameters per task. We additionally demonstrate substantial performance improvements in few-shot domain generalization across a variety of tasks. Our code is publicly available in~\url{https://github.com/rabeehk/hyperformer}.
\end{abstract}

\section{Introduction}
\begin{figure}[t]
\centerline{
\includegraphics[trim={0 0 0.42cm 0},clip, width=0.5\textwidth]{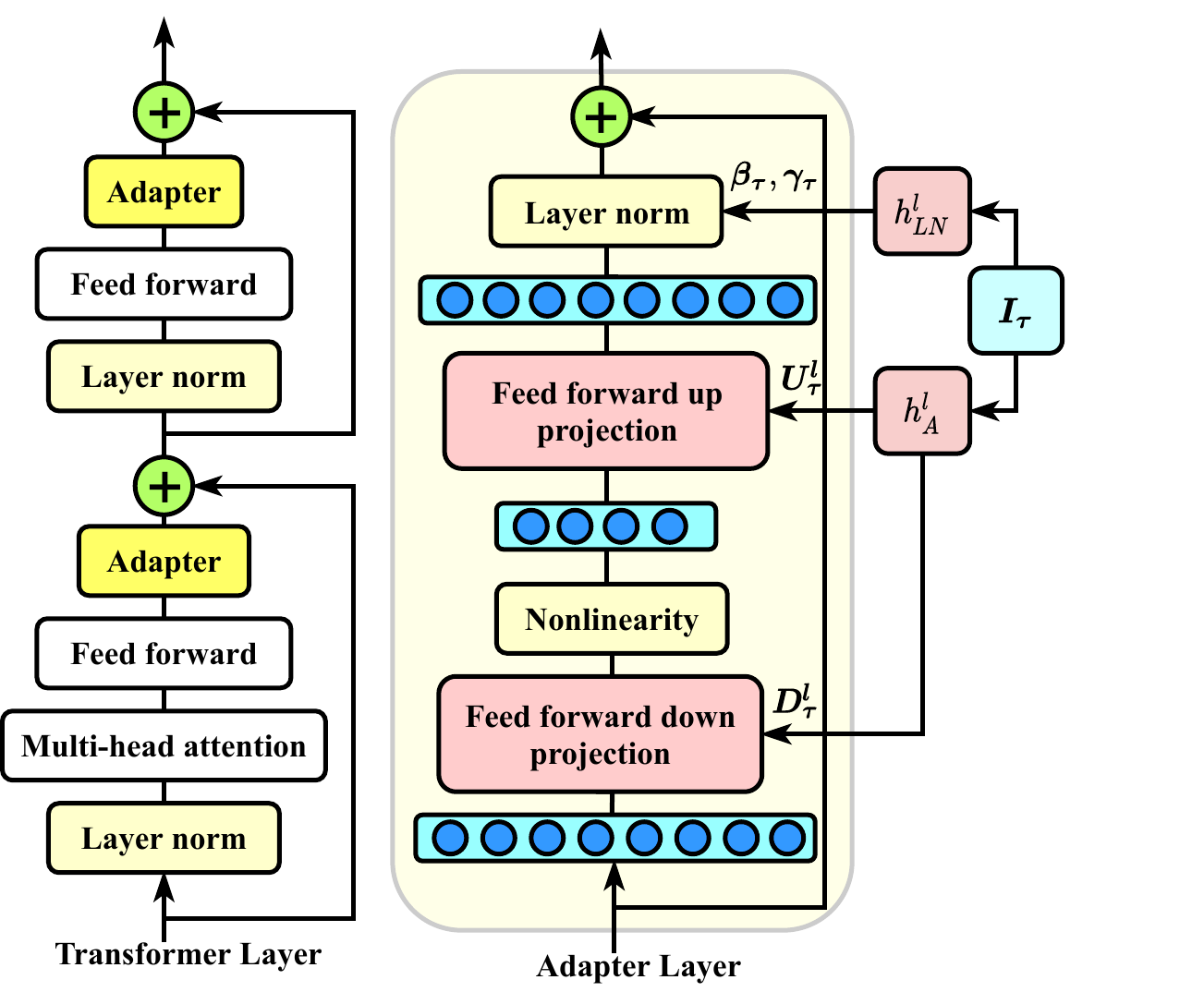}}
\vspace{-0.6em}
\caption{
Left: Adapter integration in the T5 model.  Right: Our \method adapter architecture. Following~\citet{houlsby2019parameter}, we include adapter modules after the two feed-forward layers. The Adapter hypernetwork $h_{A}^l$ produces the weights ($\bm{U_{\tau}^l}$ and $\bm{D_{\tau}^l}$) for task-specific adapter modules conditioned on an input task embedding $\bm{I_{\tau}}$. Similarly, the layer normalization hypernetwork $h_{LN}^l$ generates the conditional layer normalization parameters ($\bm{\beta_{\tau}}$ and $\bm{\gamma_{\tau}}$). During training, we only update layer normalizations in T5, hypernetworks, and task embeddings. The compact \methodefficient shares the same hypernetworks across all layers and tasks and computes the task embedding based on task, layer id, and position of the adapter module (\textsection\ref{sec:compact_model}).} \vspace{-1.5em}
\label{fig:our_method}
\end{figure} 

Transfer learning from pretrained large-scale language models yields state-of-the-art results in a variety of tasks~\citep{devlin2019bert, radford2018improving, liu2019roberta}. As a highly expressive and abstract framework,~\citet{raffel2019exploring} explored the landscape of transfer learning by converting text-based natural language processing (NLP) problems into a sequence-to-sequence format to train a unified model on several tasks simultaneously.
Multi-task learning with pretrained language models \cite{Ruder2017} is appealing for multiple reasons: 1) Training individual models per task results in higher computational costs, which hinders deployment and maintenance. These costs are substantially reduced by training a single model. 2) Fine-tuning the model across multiple tasks allows sharing information between the different tasks and positive transfer to other related tasks. Specifically, when target datasets have limited training data, multi-task learning improves the performance compared to individually trained models~\citep{liu2019multi, ratner2018snorkel}. However, multi-task fine-tuning can result in models underperforming on high-resource tasks due to constrained capacity~\cite{arivazhagan2019massively, mccann2018natural}. An additional issue with multi-task fine-tuning is the potential for \emph{task interference} or \emph{negative transfer}, where achieving good performance on one task can hinder performance on another~\citep{wang2019characterizing}.

As an alternative to fine-tuning \cite{Howard2018ulmfit}, adapter layers~\citep{houlsby2019parameter} insert a small number of additional parameters per task into the model. During fine-tuning, only the adapter modules, layer normalizations, and parameters of the final classification layer are updated, while the original pretrained model parameters remain frozen. Such task-specific adapters eliminate negative task interference by encapsulating task-specific information \cite{pfeifferetal2020adapterhub}. However, so far there has not been an effective and parameter-efficient way to share information across multiple adapters to enable positive transfer to low-resource and related tasks. 

To address this problem and to enable sharing information across tasks while reaping the benefits of adapter layers, as depicted in Figure~\ref{fig:our_method}, we propose \methodefficient, which employs a compact hypernetwork \cite{Ha2017hypernetworks,Oswald2020continual} shared across tasks and layers. The hypernetwork learns to generate task and layer-specific adapter parameters, conditioned on task and layer id embeddings. The hypernetwork is jointly learned between all tasks and is thus able to share information across them, while negative interference is minimized by generating separate adapter layers for each task.  For each new task, our model only requires learning an additional task embedding, reducing the number of trained parameters.

We use the encoder-decoder T5 model~\citep{raffel2019exploring} as the underlying model for our experiments and evaluate on the standard GLUE benchmark~\cite{wang2018glue}. We achieve strong gains over both the \basebase model as well as adapters~\citep{houlsby2019parameter}. To our knowledge, this is the first time that adapters have been successfully integrated into a state-of-the-art encoder-decoder model beyond machine translation \cite{bapnafirat-2019-simple}, demonstrating that our method effectively balances sharing information across tasks while minimizing negative transfer.

In summary, we make the following contributions: 
(1) We propose a parameter-efficient method for multi-task fine-tuning based on hypernetworks and adapter layers. (2) We demonstrate that our method scales more efficiently than prior work. (3) We provide empirical results on GLUE demonstrating the effectiveness of the proposed method on multi-task learning. (4) We perform extensive few-shot domain transfer experiments, which reveal that the captured shared knowledge can positively transfer to unseen in-domain tasks.
We release our code to facilitate future work.

\section{\method}
In this section, we present our \method model, which integrates \textbf{hyper}network-based adapter layers into a multi-task trans\textbf{former} model. In \textsection\ref{sec:compact_model}, we introduce a parameter-efficient variant of this model, called \methodefficient\!. 

\paragraph{Problem formulation:} We consider a general multi-task learning problem, where we are given the data from a set of tasks $\{\mathcal{D}_{\tau}\}_{\tau=1}^{T}$, where $T$ is the total number of tasks and $\mathcal{D}_{\tau} = \{(\bm{x_{\tau}^i}, y_{\tau}^i)\}_{i=1}^{N_{\tau}}$  shows the training data for $\tau$-th task with $N_{\tau}$ samples. We assume we are also given a large-scale pretrained language model $f_{\bm{\theta}}(.)$ parameterized by $\bm{\theta}$ that computes the output for input $\bm{x_{\tau}^i}$. Standard multi-task fine-tuning minimizes the following loss on the training set:
\myspace
\begin{align}
\hspace{-0.5em}
\mathcal{L}(\bm{\theta}, \{\mathcal{D}_{\tau}\}_{\tau=1}^{T})=\sum_{\tau=1}^{T}\sum_{~(\bm{x_{\tau}^i}, y_{\tau}^i)\in \mathcal{D}_{\tau}}\hspace{-1em} w_{\tau}l\left(f_{\bm{\theta}}(\bm{x_{\tau}^i}), y_{\tau}^i\right),
\eqend
\end{align}
where $l$ is typically the cross-entropy loss, and $w_{\tau}$ shows the sampling weight for $\tau$-th task. Our goal is to finetune the pretrained model in a multi-task learning setup efficiently, while allowing sharing information across tasks and at the same time, enabling the model to adapt to each individual task. 

The key idea of our approach, depicted in Figure ~\ref{fig:our_method}, is to learn a parametric task embedding $\{\bm{I_{\tau}}\}_{\tau=1}^{T}$ for each task, and then feed these task embeddings to hypernetworks parameterized by $\bm{\nu}$ that generate the task-specific adapter layers~\citep{houlsby2019parameter}. We insert adapter modules within the layers of a pretrained model, making the final model of $\mathcal{X}_{\bm{\nu}}(\bm{x_{\tau}^i}, \bm{\theta}, \bm{I_{\tau}})$ parameterized by $\bm{\nu}$ that computes the output for input $\bm{x_{\tau}^i}$. During training, we only train hypernetwork parameters $\bm{\nu}$, task embeddings $\{\bm{I_{\tau}}\}_{\tau=1}^{T}$, and layer normalizations in $f_{\theta}(.)$, while the rest of the pretrained model parameters $\bm{\theta}$ are fixed:
\myspace
\begin{align}
&\mathcal{L}(\bm{\nu}, \{\bm{I_{\tau}}\}_{i=1}^T, \{\mathcal{D}_{\tau}\}_{\tau=1}^{T})= \nonumber \\
&\sum_{\tau=1}^{T}\sum_{(\bm{x_{\tau}^i}, y_{\tau}^i)\in \mathcal{D}_{\tau}}w_{\tau}l\left(\mathcal{X}_{\bm{\nu}}( \bm{x_{\tau}^i}, \bm{\theta}, \bm{I_{\tau}}), y_{\tau}^i\right), 
\eqend
\end{align}
The hypernetworks capture the shared information across tasks in a multi-task learning model enabling positive transfer between related domains and transferable
tasks, while adapters are reducing negative interference, encapsulating task-specific information.

\paragraph{Base model:} All of our models are built on top of the state-of-the-art T5 transformer model~\citep{raffel2019exploring}. This model frames text-based language tasks as sequence-to-sequence problems. T5 consists of an encoder-decoder Transformer~\citep{vaswani2017attention} with minor modifications~\citep{raffel2019exploring}. The model is trained simultaneously on multiple tasks, obtaining state-of-the-art performance across a diverse set of tasks. We use the T5 framework as it enables training a universal model that interfaces with many language tasks.
Our model has three main components: 1) task conditional adapter layers; 2) task conditional layer normalizations; and 3) hypernetworks that generate task-specific parameters.  We next describe these components.

\vspace{-0.2em}
\subsection{Task Conditional Adapter Layers}\label{sec:task-adapter}
Prior work has shown that fine-tuning all parameters of the model can result in a sub-optimal solution, particularly for resource-limited datasets~\citep{peters2019tune}. As an alternative to fine-tuning all the model's parameters, prior work \cite{houlsby2019parameter,rebuffi2018efficient, stickland2019bert} inserted small modules called \emph{adapter layers} within layers of a pretrained model, as shown in Figure~\ref{fig:our_method}. Adapters introduce no change to the structure or parameters of the original model.

In this work, we propose conditional adapter modules, in which we generate the adapters weights based on input task embeddings using shared hypernetworks~\citep{Ha2017hypernetworks}, which capture information across tasks that can be used to positively transfer to other relevant tasks.

Each layer of a transformer model consists of an attention block and a feed-forward block, each followed by a skip connection. Following~\citet{houlsby2019parameter}, as depicted in Figure~\ref{fig:our_method}, we introduce a conditional adapter layer after each block before the skip connection. 
%
The conditional adapter layer $A_{\tau}^l$ for layer $l$ consists of a down-projection, $\bm{D_{\tau}^l}\in\mathbb{R}^{h\times d}$, GeLU non-linearity~\citep{hendrycks2016gaussian}, and up-projection $\bm{U^l_{\tau}} \in\mathbb{R}^{d \times h}$, where $h$ is the input dimension, and $d$ is the bottleneck dimension for the adapter layer, mathematically defined as: 
\myspace
\begin{align}
A^l_{\tau}(\bm{x}) = {LN}^{l}_{\tau}\left(\bm{U^l_{\tau}}(\text{GeLU}(\bm{D^l_{\tau}}(\bm{x})))\right) + \bm{x}, 
\eqend
\end{align}
where $\bm{x}$ is the input hidden state and ${LN}^{l}_{\tau}$ is the conditional layer norm defined in the next section. We generate adapter weights ($\bm{U^l_{\tau}}$, $\bm{D^l_{\tau}}$) through a hypernetwork described in \textsection\ref{sec:hypernetworks}.

\subsection{Task Conditional Layer Normalization}
Conventional layer normalization~\cite{ba2016layer} is defined as: 
\myspace
\begin{align}
{LN}^{l}_{\tau}(\bm{x_{\tau}^i}) = \bm{\gamma^l_{\tau}} \odot  \frac{\bm{x_{\tau}^i}-\bm{\mu_{\tau}}}{\bm{\sigma_{\tau}}} + \bm{\beta^l_{\tau}},
\eqend
\end{align} 
where  $\odot$ is the element-wise multiplication between two vectors, and $\bm{\gamma^l_{\tau}}$ and $\bm{\beta^l_{\tau}}$ are learnable parameters with the same dimension as $\bm{x_{\tau}^i}$. Values of $\bm{\mu_{\tau}}$ and $\bm{\sigma_{\tau}}$ show the mean and standard deviation of training data for the $\tau$-th task.

To allow the layer normalization inside adapters to adapt to each task, inspired by~\citet{perez2018film, de2017modulating}, we generate $\bm{\gamma^l_{\tau}}$, $\bm{\beta^l_{\tau}}$ via a hypernetwork as a function of task embeddings (\textsection\ref{sec:hypernetworks}).

\subsection{Task Conditioned Hypernetworks}\label{sec:hypernetworks}
In order to have a model that can share information while being able to adapt to each individual task, we generate the parameters of task conditional adapter layers and layer normalization using hypernetworks. A hypernetwork is a network that generates the weights of another network~\citep{Ha2017hypernetworks}.

The hypernetworks capture the shared information, while the generated task conditional adapters and layer normalization allow the model to adapt to each individual task to reduce negative task interference. 

\paragraph{Learned task embedding:} We first compute a task embedding $\bm{I_{\tau}} \in \mathbb{R}^t$ for each individual task using a task projector network $h_I(.)$, which is a  multi-layer perceptron consisting of two feed-forward layers and a ReLU non-linearity:
\myspace
\begin{align}
\bm{I_{\tau}} = h_{I}(\bm{z_{\tau}}), \label{eqn:task_projector}
\eqend
\end{align} 
where $\bm{z_{\tau}}\in\mathbb{R}^{t'}$ can be a learnable parameter or any pretrained task features~\citep{vuetal2020exploring}, and the task projector network $h_I(.)$ learns a suitable compressed task embedding from input task features. In this work, we consider a parametric $\bm{z_{\tau}}$ to allow end-to-end training which is convenient in practice.\footnote{We ran some pilot experiments with pretrained task embeddings~\citep{vuetal2020exploring}, but did not observe extra benefits.}
\paragraph{Removing task prefixes:} The T5 model prepends task-specific prefixes to the input sequence for conditioning. For instance, when training on CoLA~\citep{warstadt-etal2019neural}, \emph{cola sentence:} is prepended to each sample. Instead, we remove task prefixes and use task embeddings for conditioning.

\paragraph{Task conditioned hypernetworks:} We consider simple linear layers as hypernetworks that are functions of input task embeddings $\bm{I_{\tau}}$. We introduce these hypernetworks in each layer of the transformer. We define hypernetwork $h_A^l(.)$ that generates task conditional adapter weights ($\bm{U^l_{\tau}}$, $\bm{D^l_{\tau}}$):
\myspace
\begin{align}
(\bm{U^l_{\tau}}, \bm{D^l_{\tau}}) := h_A^l(\bm{I_{\tau}}) = \left( \bm{W^{U^l}},  \bm{W^{D^l}}\right) \bm{I_{\tau}}, 
\eqend
\end{align} where $\bm{W^{U^l}} \in \mathbb{R}^{(d\times h)\times t}$ and $\bm{W^{D^l}} \in \mathbb{R}^{(h\times d)\times t}$ are the respective hypernetwork parameters. 
We additionally define the hypernetwork $h_{LN}^{l}(.)$ that computes the layer normalization parameters:
\myspace
\begin{align}
(\bm{\gamma_{\tau}^l}, \bm{\beta_{\tau}^l}) := h_{LN}^{l}(\bm{I_{\tau}}) = \left(\bm{W^{\gamma^l}}, \bm{W^{\beta^l}}\right) \bm{I_{\tau}}, 
\eqend
\end{align}
where $\bm{W^{\gamma^l}} \in \mathbb{R}^{h \times t}$ and $\bm{W^{\beta^l}} \in \mathbb{R}^{h \times t}$. 

\subsection{\methodefficient}\label{sec:compact_model}  
A downside of introducing a separate hypernetwork in each layer of the Transformer is that it increases the overall number of parameters. We, therefore, propose to share hypernetworks across transformer layers. By having a shared hypernetwork that is reusable, this strategy results in a substantial reduction in the number of parameters. However, reapplying the same hypernetwork across all the layers introduces weight sharing across target parameters, which may not be desirable. To allow for a flexible parameterization of task conditional adapters/layer normalization, for a transformer of $L$ layers, we introduce a set of \emph{layer id} embeddings $\mathcal{I}=\{\bm{l_i} \}_{i=1}^L$, and \emph{adapter position} embeddings $\mathcal{P}=\{\bm{p_j}\}_{j=1}^2$, which specify the position of adapter layers in each transformer block (after the attention layer or feed-forward layer), which are used as additional inputs to the hypernetworks. For simplicity, we consider $\bm{l_i}\in\mathbb{R}^t$, $\bm{p_j}\in\mathbb{R}^t$, and $\bm{z_{\tau}}\in\mathbb{R}^t$. We feed a concatenation of $(\bm{z_{\tau}}, \bm{l_i}, \bm{p_j})$ to a similar task projector network $h'_{I}$ as in Eq.~\eqref{eqn:task_projector}: 
\myspace
\begin{align}
\bm{I_{\tau}} = h'_{I}(\bm{z_{\tau}}, \bm{l_i}, \bm{p_j}), \label{eqn:compact_task_projector}
\eqend
\end{align}
which is then followed by a shared layer normalization to compute final task embeddings $\bm{I_\tau}\in\mathbb{R}^{t}$ to the hypernetwork. This way, the hypernetwork is able to produce distinct weights for each task, adapter position, and layer of a transformer. Furthermore, layer id and adapter position embeddings are parameters that are learned via back-propagation, allowing us to train the whole model end-to-end conveniently. 
\vspace{-0.1em}

\begin{table*}[ht!]
\centering 
\begin{adjustbox}{max width=\textwidth}
\begin{tabular}{l|ll|llllllll|l}
\toprule 
\textbf{Model} & \pbox{3cm}{\textbf{\#Total}\\ \textbf{params}} & \pbox{3cm}{\textbf{\#Trained} \\ \textbf{params /}\\ \textbf{per task\vspace{0.1em}}} & \textbf{CoLA} &    \textbf{SST-2} &   \textbf{MRPC} &   \textbf{QQP} &     \textbf{STS-B} & \textbf{MNLI}  &    \textbf{QNLI} & \textbf{RTE} &   \textbf{Avg} \\
\toprule 
\rowcolor{gray!20}\multicolumn{12}{c}{\it \textbf{Single-Task Training}}\\
\midrule 
\basesmall &  $8.0\times$ &  $100\%$  &  \textbf{46.81} &        \textbf{90.47} &        \textbf{86.21/90.67} &       \textbf{91.02/87.96} &            \textbf{89.11/88.70} &        \textbf{82.09} &        \textbf{90.21} &       \textbf{59.42} & \textbf{82.06}\\
\adaptersmall~\ding{163} & $1+8\times0.01$  & \bm{${0.74\%}$}  &40.12 &        89.44 &        85.22/89.29 &       90.04/86.68 &            83.93/83.62 &        81.58 &        89.11 &        55.80 &  79.53 \\
\midrule 
\basebase &  $8.0\times$ &   $100\%$ &  54.85 & 92.19 & \textbf{88.18/91.61} & \textbf{91.46/88.61} &  \textbf{89.55/89.41} & \textbf{86.49}& 91.60& 67.39 & 84.67 \\
\adapterbase~\ding{163} &  $1+8\times0.01$ & \bm{${0.87\%}$}  &        \textbf{59.49} &        \textbf{93.46} & 88.18/91.55 &       90.94/88.01 &            87.44/87.18 &        86.38 &        \textbf{92.26} &       \textbf{68.84} & \textbf{84.88}  \\
\midrule
\rowcolor{gray!20}\multicolumn{12}{c}{\it \textbf{Multi-Task Training}}\\
\midrule 
\basesmall~\ding{171} &  $1.0\times$ &   $12.5\%$ &        50.67 &        \textbf{91.39} &        84.73/88.89 &       \textbf{89.53/86.31} &             \textbf{88.70/88.27} &        81.04 &        89.67 &       59.42  & 81.69  \\
\madaptersmall &  $1.05 \times$ &  $0.68\%$ &    39.87 &        90.01 &        88.67/91.81 &       88.51/84.77 &            88.15/87.89 &        79.95 &        89.60  &       60.14 &80.85  \\
\methodsmall &  $1.45 \times$   & $5.80\%$ &  47.64 &        \textbf{91.39} &        \textbf{90.15/92.96} &      88.68/85.08 &            87.49/86.96 &        \textbf{81.24} &        \textbf{90.39} &       65.22 & 82.47  \\ 
\methodefficientsmall & $1.04 \times$ & \bm{${0.50\%}$}&     \textbf{53.96} &        90.59 &        84.24/88.81 &       88.44/84.46 &            87.73/87.26 &        80.69 &        \textbf{90.39} &       \textbf{71.01} &  \textbf{82.51}  \\ 
\midrule 
\basebase~\ding{171} &  $1.0\times$ &   $12.5\%$ &         54.88 &        92.54 &        90.15/93.01 &       \textbf{91.13/88.07} &            88.84/88.53 &        85.66 &        92.04 &       75.36 &  85.47\\
\madapterbase &  $1.07\times$ &  $0.82\%$  &        61.53 &           93.00 &        90.15/92.91 &       90.47/87.26 &            89.86/89.44 &        86.09 &        \textbf{93.17} &       70.29 & 85.83 \\ 
\methodbase & $1.54\times$& $6.86\%$ &       61.32 &         93.80 &        \textbf{90.64/93.33} &       90.13/87.18 &            89.55/89.03 &        \textbf{86.33} &        92.79 &       \textbf{78.26} & \textbf{86.58}   \\
\methodefficientbase & $1.02\times$ & \bm{${0.29\%}$} &    \textbf{63.73} &        \textbf{94.03} &        89.66/92.63 &       90.28/87.20 &               \textbf{90.00/89.66} &        85.74 &        93.02 &       75.36 & 86.48  \\ 
\bottomrule
\end{tabular}
\end{adjustbox} \vspace{-0.5em}
\caption{Performance of all models on the GLUE tasks. For each method, we report the total number of parameters across all tasks and the number of parameters that are trained for each task as a multiple and proportion respectively of the corresponding single-task T5 model.
For MNLI, we report accuracy on the matched validation set. For MRPC and QQP, we report accuracy and F1. For STS-B, we report Pearson and Spearman correlation coefficients. For CoLA, we report Matthews correlation. For all other tasks, we report accuracy. 
\madapter refers to our proposed variant of adapters with shared layer normalizations.
Our \methodefficient\! obtains a better score on average compared to full fine-tuning and \madapter\!\!, while being more parameter-efficient. \ding{171}: Our re-implementation of~\citet{raffel2019exploring}, \ding{163}: Applying method of~\citet{houlsby2019parameter} on T5. Bold fonts indicate the best results in each block.
\label{tab:glue_results}}\vspace{-1em}
\end{table*} 
\vspace{-0.2em}
\section{Experiments} \label{sec:experiments}
\paragraph{Datasets:} Following~\citet{raffel2019exploring}, we evaluate the performance of the models on the GLUE benchmark~\citep{wang2018glue}. This benchmark covers multiple tasks of paraphrase detection (MRPC, QQP), sentiment classification (SST-2), natural language
inference (MNLI, RTE, QNLI), and linguistic acceptability (CoLA).\footnote{
Following~\citet{raffel2019exploring, devlin2019bert}, as a common practice, due to the adversarial nature of WNLI with respect to the training set, we do not experiment with WNLI.
}
The original test sets are not publicly available, and following~\citet{zhang2020revisiting}, for datasets fewer than 10K samples (RTE, MRPC, STS-B, CoLA),  we divide the original validation set in half, using one half for validation and the other for the test. For the other larger datasets, we split 1k samples from the training set as our validation data and test on the original validation set.
\paragraph{Experimental details:} 
We use the HuggingFace implementation~\citep{wolf-etal2020transformers} of the T5 model~\cite{raffel2019exploring}. We fine-tune all models with a constant learning rate of $0.0003$ and following~\citet{raffel2019exploring}, we use $2^{18}=262144$ steps in all experiments. We save a checkpoint every $1000$ steps for all models (see also \textsection\ref{sec:appendix}). \citet{raffel2019exploring} report the results based on the best checkpoint for each task independently. In contrast, we focus on the more realistic setting where we report the results on a single checkpoint with the highest average validation performance across all tasks. The hyperparameters are selected in the same manner.
In contrast to prior work~\citep{houlsby2019parameter}, we do not learn a separate output layer for each task but instead share a frozen output layer for all the tasks, which makes our setting more parameter-efficient than prior work and is an advantage of multi-task learning with encoder-decoder models.\footnote{According to our initial experiments, fine-tuning the final output layer did not improve performance for adapter-based methods.} 

\paragraph{Baselines:} We compare to the strong adapter baseline~\citep{houlsby2019parameter}. Following~\citet{houlsby2019parameter}, we add adapters modules for each task after the two feed-forward modules in each transformer block of the T5 model. As suggested in~\citet{houlsby2019parameter}, we train the layer normalization parameters inside the T5 model, per task.
We refer to this method as \emph{Adapters}. We additionally propose a variant of this model, in which we share all layer normalization parameters (T5 and adapters) across all tasks. We refer to this model as \madapteritalic. We compare our models to the state-of-the-art T5 model, in which we fine-tune all parameters of the model on all tasks. We refer to this method as \basesmall/\basebase in experiments.

\paragraph{Sampling tasks:} During training, we sample tasks with conventional temperature-based sampling with temperature $T=10$ for all methods. We sample different tasks proportional to $p_{\tau}^{1/T}$ where $p_{\tau} = \frac{N_{\tau}}{\sum_{i=1}^T N{\tau}}$ and $N_{\tau}$ is the number of training samples for the $\tau$-th task. We did not experiment with more complex sampling strategies~\citep{raffel2019exploring} or tuning of $T$.

\subsection{Results on the GLUE Benchmark}
 Table~\ref{tab:glue_results} shows the results on \glue for single-task and multi-task training. We experiment with reduction factors of $r=\{8, 16, 32\}$ for all adapter-based methods, where $r=\frac{h}{d}$. We report the results both with \basesmall\!~(6 layers and 60M parameters) and \basebase~models (12 layers and 222M parameters). 

Overall, our proposed \methodefficient\! obtains strong gains over Adapters (82.51 versus 79.53 for \basesmall~and 86.48 versus 84.88 for \basebase) while being more parameter-efficient. 

Our variant of Adapters$\dagger$, which shares layer norms across tasks, outperforms prior work \citep{houlsby2019parameter}, which does not share such information (80.85 versus 79.53 for \basesmall~and 85.83 versus 84.88 for \basebase). This demonstrates that in encoder-decoder models such as T5 more sharing of information across tasks is beneficial.

Our proposed \method obtains consistent improvement over our proposed \madapter method. We attribute this improvement to the ability to learn the shared information across tasks through our hypernetworks. Interestingly, \methodefficient\! obtains similar performance as \method while being more than an order of magnitude more parameter-efficient. Adapter modules thus seem to be similar enough so that much of their information can be modeled by a single, appropriately conditioned network.

Compared to single-task fine-tuning of all parameters, our methods on average improve the results by 0.45 for \basesmall~and 1.81 for \basebase~with substantial improvement on low-resource datasets like CoLA (63.73 versus 54.85) and RTE (75.36 versus 67.39) due to shared hypernetworks that capture the shared information and enable positive transfer effects. 

We also report the total number of parameters and trainable parameters for all methods in Table~\ref{tab:glue_results}. For adapter-based methods, the number of parameters varies based on the adapter size (we report all numbers with $r=32$). The multiple in terms of the number of parameters of \methodefficientbase with regard to \basebase is $1.02\times$ with only $0.29\%$ trainable parameters per task. Note that by keeping the output layer frozen for \adaptersmall and \adapterbase, they require $5.51 \times $ and $2.53 \times$ fewer parameters respectively compared to a direct application of prior work \citep{houlsby2019parameter}. Despite using more efficient baselines, compared to \adapterbase, \methodefficientbase requires $3 \times$ fewer trainable parameters. 

\subsection{Few-shot Domain Transfer}\label{sec:fewshot}
Finally, we assess how well a trained \method can generalize to new tasks.  We evaluate performance on 5 tasks and 7 datasets. In particular, we consider 1) the natural language inference (NLI) datasets SciTail~\citep{khot2018scitail}, and CB~\citep{de2019commitmentbank} from SuperGLUE~\citep{wang2019superglue} 2) the question answering (QA) dataset BoolQ~\citep{clark2019boolq}; 3) the sentiment analysis datasets IMDB~\citep{maas-EtAl:2011:ACL-HLT2011} and Yelp Polarity~\citep{zhang2015character}; and 4) the paraphrase detection dataset PAWS~\citep{pawsbaldridge2019}; 5) the question classification dataset TREC~\citep{li2002learning}.

\begin{table}[thp!] 
    \centering
     \resizebox{0.5\textwidth}{!}{
    \begin{tabular}{lllll}
    \toprule 
    \textbf{\small Dataset} & \rotatebox{45}{\textbf{\small \# Samples}} & \rotatebox{50}{\textbf{\small \basebase}} & \rotatebox{50}{\textbf{\small \madapterbase}} & \rotatebox{50}{\textbf{\small \methodefficientbase}} \\ 
    \toprule 
    \rowcolor{gray!20}\multicolumn{5}{c}{\it \textbf{Natural Language Inference}}\\
   \toprule 
  
    \multirow{6}{*}{SciTail} 
             & 4 &  79.60\std{3.3} & 79.54\std{2.8}& \textbf{82.00}\std{4.9} \\
             & 16 & 80.03\std{2.3} & 83.25\std{1.7}& \textbf{86.55}\std{1.4}  \\ 
             & 32 &81.97\std{1.3} &  85.06\std{1.1}& \textbf{85.85}\std{1.4} \\
             & 100 & 84.04\std{0.7} & 88.22\std{1.3} & \textbf{88.52}\std{0.7}\\ 
             & 500 & 88.07\std{0.7} &  91.27\std{0.8} & \textbf{91.44}\std{0.6}\\ 
             & 1000 & 88.77\std{1.0} &  91.75\std{0.8} & \textbf{92.34}\std{0.5}\\
             & 2000 & 91.01\std{1.0} &  92.72\std{0.5} &  \textbf{93.40}\std{0.2}\\
    \midrule 
    \multirow{6}{*}{CB} & 4 & 57.78\std{10.9}& 51.11\std{9.2}& \textbf{60.74}\std{16.66}  \\
             & 16 & \textbf{77.04}\std{7.2} & 74.81\std{5.4} & 76.29\std{4.45}\\
             & 32 & 80.0\std{7.6}  &  74.81\std{5.9} & \textbf{81.48}\std{6.2}\\
             & 100 & 85.93\std{5.4}&  80.74\std{7.6} & \textbf{87.41}\std{2.96} \\ 
             & 250 &85.19\std{4.7} & 86.67\std{5.0} & \textbf{89.63}\std{4.32}\\

    \midrule 
      \rowcolor{gray!20}\multicolumn{5}{c}{\it \textbf{Question Classification}}\\
      \midrule 
     \multirow{6}{*}{TREC} & 4 &28.11\std{5.9} & 23.61\std{7.7}& \textbf{28.85}\std{6.9}\\ 
      & 16 & 40.08\std{12.6}& 43.45\std{14.0}& \textbf{49.40}\std{9.5} \\ 
      & 32&62.49\std{6.2} & 59.6\std{7.0}& \textbf{68.94}\std{7.5}\\ 
    & 100&87.79\std{0.7} & 78.07\std{3.8}& \textbf{88.42}\std{1.7} \\ 
    & 500& 93.57\std{1.3} & 93.65\std{1.7}& \textbf{94.78}\std{1.4} \\ 
    & 1000&95.5\std{0.9}  & 96.06\std{0.4}&   \textbf{96.72}\std{1.3}\\ 
    & 2000&96.87\std{1.3} &  \textbf{97.03}\std{0.7}& 96.92\std{0.9}\\ 
   
    \midrule 
    \rowcolor{gray!20}\multicolumn{5}{c}{\it\textbf{Question Answering}}\\
    \midrule  
     \multirow{6}{*}{BoolQ} & 4 &  50.49\std{11.1} & \textbf{53.48}\std{2.8} & 48.03\std{4.8}  \\
             & 16 &  \textbf{56.50}\std{7.1} &  51.37\std{6.5} & 50.21\std{7.9}\\ 
             & 32 & \textbf{58.43}\std{4.9} & 54.52\std{5.1}& 58.37\std{3.7}\\
             & 100 &  60.10\std{2.4} & 58.60\std{1.6}& \textbf{62.03}\std{2.0}\\ 
             & 500 & 66.49\std{1.2} &66.72\std{0.7}&  \textbf{70.04}\std{1.4}\\ 
             & 1000 & 69.01\std{1.1} &70.21\std{1.3}& \textbf{72.35}\std{1.7}\\
             & 2000 & 71.58\std{0.8} &73.60\std{0.8}& \textbf{74.94}\std{0.6}\\
    \midrule 
      \rowcolor{gray!20}\multicolumn{5}{c}{\it\textbf{Sentiment Analysis}} \\
     \midrule 
       \multirow{6}{*}{IMDB} & 4 & 77.23\std{3.0} & 81.55\std{1.9}& \textbf{81.77}\std{1.8}    \\
             & 16 &  82.74\std{1.7} & 82.54 \std{1.0}& \textbf{84.06}\std{0.7}\\ 
             & 32 &  83.42\std{1.0} & 83.39 \std{0.8}& \textbf{84.64}\std{0.4}\\
             & 100 &84.58\std{0.6} & 83.35 \std{0.8}& \textbf{84.74}\std{0.4} \\ 
             & 500 & 84.99\std{0.3}   &  85.37\std{0.5} &  \textbf{86.00}\std{0.2}\\ 
             & 1000 &85.50\std{0.1} &86.27\std{0.4} & \textbf{86.37} \std{0.4}\\
             & 2000 & 86.01\std{0.2} &86.57\std{0.2} & \textbf{86.60}\std{0.1}\\
            \midrule 
          \multirow{6}{*}{Yelp polarity} & 4 & 76.85\std{14.3} & 81.37\std{13.1} & \textbf{90.25}\std{1.0}\\
             & 16 & 87.84\std{1.5} &\textbf{91.08}\std{0.2} &90.36\std{1.2}\\ 
             & 32 &  89.22\std{0.7} &  91.09\std{0.5}& \textbf{91.15}\std{0.5}\\
             & 100 & 90.19\std{0.7} & 90.15\std{0.7}& \textbf{91.06}\std{0.6}\\ 
             & 500 & 90.92\std{0.2} & 91.52\std{0.2} &\textbf{92.09}\std{0.4} \\ 
             & 1000 & 91.32\std{0.2} &92.26\std{0.6} &\textbf{92.50}\std{0.2}\\
             & 2000 & 91.68\std{0.1} &92.36\std{0.4} & \textbf{92.70}\std{0.1} \\
        \midrule 
        \rowcolor{gray!20}\multicolumn{5}{c}{\it \textbf{Paraphrase Detection}} \\
         \midrule 
           \multirow{6}{*}{PAWS} & 4 &53.89\std{3.6}&\textbf{55.69}\std{9.0} & 55.58\std{7.5} \\
             & 16 & 54.18\std{1.0}& 63.38\std{5.3} &\textbf{72.71}\std{1.1} \\ 
             & 32 &55.23\std{3.2} & 68.78\std{1.5} & \textbf{73.39}\std{2.1} \\
             & 100 &  71.51\std{2.4}& 73.82\std{1.6} & \textbf{78.24}\std{2.1}\\ 
             & 500 & 82.81\std{1.0} & 85.36\std{0.6} &  \textbf{86.3}\std{1.1}\\ 
             & 1000 &  85.67\std{0.7}& 87.89\std{0.6} &  \textbf{89.12}\std{0.5}\\
             & 2000 &  88.33\std{0.6} &  90.41\std{0.6} &\textbf{90.87}\std{0.3}\\
    \bottomrule
    \end{tabular}}
    \vspace{-1ex}
    \caption{Few-shot domain transfer results of the models trained on GLUE averaged across 5 seeds. We compute accuracy for all datasets.} 
    \label{tab:transfer_results} \vspace{-1em}
\end{table}

For CB and BoolQ, since test sets are not available, we divide the validation sets in half, using one half for validation and the other for testing. For Yelp polarity, TREC, and IMDB, since validation sets are not available, we similarly divide the test sets to form validation sets. For the rest, we report on the original test sets.

We consider the models trained on GLUE reported in Table~\ref{tab:glue_results} and evaluate them on the test set after the few-shot fine-tuning on each target training data. For \madapter and our method, we use the adapter and the task embedding respectively trained on the most similar GLUE task for initialization, i.e. MNLI for NLI, QNLI for QA, SST-2 for sentiment analysis, and QQP for paraphrase detection. Following prior evidence of positive transfer from NLI to other tasks~\cite{conneau2018senteval, yin2020universal, phang2018sentence}, we initialize the out-of-domain TREC from MNLI. 
We show the results of full fine-tuning of all model's parameters, \madapter, and \methodefficient\footnote{We finetune hypernetworks and task embeddings parameters. We also tried only fine-tuning the task embedding but found that this achieves lower performance in the few-shot setting and comparable performance with more samples.} in Table~\ref{tab:transfer_results}. Our method significantly surpasses the baselines on the majority of settings. 

\subsection{Low-resource Fine-tuning} \label{sec:lowresource}

\begin{figure}[h!]
\centerline{
\includegraphics[width=0.35\textwidth]{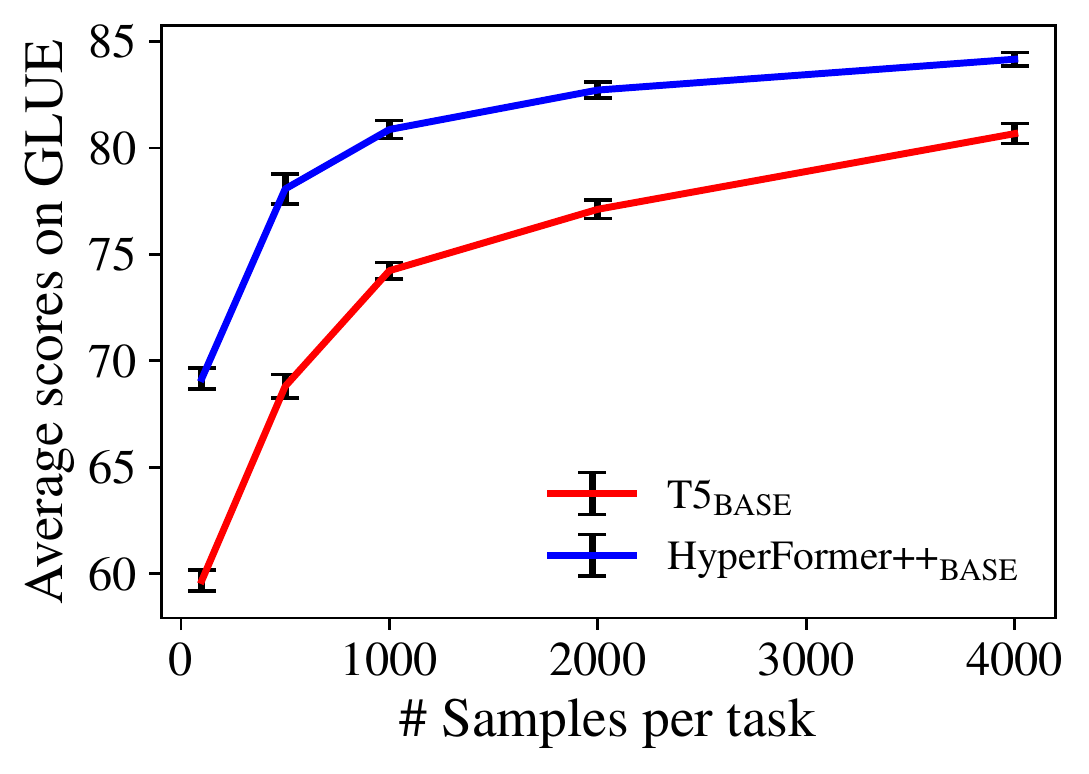}}
\caption{Results on GLUE for the various number of training samples per task $(100, 500, 1000, 2000, 4000)$. We show mean and standard deviation across 5 seeds.}
\label{fig:low_resource} 
\end{figure}

Given that our model \methodefficientbase has substantially fewer trainable parameters than \basebase, we investigate whether it generalizes better in a low-resource setting. We subsample each individual task in GLUE for varying training sizes.  We train the models for 15,000 steps, which we found to be sufficient to allow them to converge. Figure~\ref{fig:low_resource} shows the results. \methodefficientbase~substantially improves results with limited training data, indicating more effective fine-tuning in this regime.

\section{Analysis}
\subsection{Parameter Efficiency} \label{sec:parameters} 
In this section, we compare the number of parameters of \methodefficient\! with \adapter.

\paragraph{\adapter parameters:} The standard setting \cite{houlsby2019parameter} employs two adapters per layer for each task. Each adapter layer has $2hd$ parameters for projection matrices ($\bm{U_{\tau}^l}$ and $\bm{D_{\tau}^l}$) and $2h$ parameters for the layer normalization. The total number of parameters for \adapter for $L$ Transformer layers in both an encoder and a decoder across $T$ tasks is, therefore, $4TL(2hd+2h)$, which scales linearly with the number of tasks times the number of layers.

\paragraph{\methodefficient\! parameters:} Our approach learns a task feature embedding per task, consisting of $Tt$ parameters. We additionally employ layer id and adapter position embeddings in the encoder and decoder, which require $2(2+L)t$ parameters, with a fixed embedding size of $t$ for all these feature embeddings. 
We consider a separate task projector networks $h'_I$ for encoder and decoder, which is in both cases a two-layer MLP, consisting of a total of $2(3te+et)$ parameters, where $e=128$ is the hidden dimension for the task-projector network. Our hypernetwork for adapters in encoder/decoder consists of $2(2thd)$ parameters and our layer normalization hypernetwork consists of $2(2th)$ parameters. In total, this results in $\underbrace{t(T+4+2L)}_{\text{Task features}}+\underbrace{8te+2t(2hd+2h)}_\text{Hypernetworks}$ parameters. The total number of parameters for hypernetworks remains constant, while the task feature parameters scale with the number of tasks or layers times $t$, where $t=64$ in our experiments. 

In settings with a large number of layers and a large number of tasks, since $t \ll 2hd{+}2h$ and $T{+}L \ll TL$, our method is much more parameter-efficient compared to \adapter. In the current setting, the term $hd$ is the largest term, 
and the factor $2TL$ for \adapter is larger than the factor $t$ for \methodefficient.

\subsection{Do Extra Parameters Make a Difference?} 
While our \methodefficient\! is more parameter-efficient than the baselines, the number of parameters of \method per task is higher compared to \madapter. To confirm that the improvements of \method are due to its capability of sharing information across tasks and not the number of parameters, as an ablation, we run the \madapter with $r=\{2, 4\}$ and choose the model performing the best on the validation set. This allows \madapter to have a higher number of parameters compared to \method. We report the results in Table~\ref{tab:adapter_results_with_more_params} and compare them with results of \method in Table~\ref{tab:glue_results}. The results demonstrate that even with an increased number of parameters, \madapter is not able to reach the performance of \method, and \method performs substantially better.

\begin{table}[t]
    \centering
    \resizebox{\columnwidth}{!}{
    \begin{tabular}{llll}
    \toprule 
    \textbf{Model} & \textbf{GLUE} &  \pbox{3cm}{\textbf{\#Total}\\ \textbf{params}} & \pbox{3cm}{\textbf{\#Trained} \\ \textbf{params/task}} \\ 
    \toprule 
   \madapter\textsubscript{SMALL}       &  80.97  & 1.83x & 10.44\% \\  \method\textsubscript{SMALL} & 82.47 &  1.45x & 5.80 \%\\  
    \midrule 
    \madapter\textsubscript{BASE}     & 85.84 & 2.02x & 12.73\% \\ 
    \method\textsubscript{BASE}  & 86.58 & 1.54x & 6.86\%\\ 
    \bottomrule
    \end{tabular}}
    \caption{Averaged test results on \glue for \method and \madapter\!\!\!, where \madapter has a higher number of parameters compared to \method.} 
    \label{tab:adapter_results_with_more_params}
\end{table}

\subsection{Impact of the Framework Components}
We investigate the impact of the components of our framework including: (1) task conditional adapter blocks; (2) task conditional layer normalization; (3) task projection network; (4) fine-tuning of layer normalizations in the T5 model; (5) task conditional layer normalization in adapter modules and fine-tuning of layer normalizations inside the T5 model. We consider our small model of Table~\ref{tab:glue_results} and train different variants of it. Table~\ref{table:ablation_results} shows the results on GLUE, demonstrating that each component of the model contributes positively to its final performance. 

\begin{table}[t]
    \centering
    \begin{tabular}{lc}
    \toprule 
    \textbf{Model variant} & \textbf{GLUE} \\ 
    \toprule 
    \methodsmall     & 82.47 \\
    $-$ Adapter blocks &  68.37 \\
    $-$ Conditional layer norm    & 79.83  \\ 
    $-$ Task projector &  81.56 \\ 
    $-$ T5 Layer norm  & 81.29 \\ 
    $-$ Conditional layer norm, T5 Layer norm&  78.92 \\ 
    \bottomrule
    \end{tabular}
    \caption{Impact when removing different components of our framework. We report the average results on GLUE.}
    \label{table:ablation_results} 
\end{table}

\subsection{Visualization of Task Embeddings}
\begin{figure}[t]
\centerline{
\includegraphics[width=0.5\textwidth]{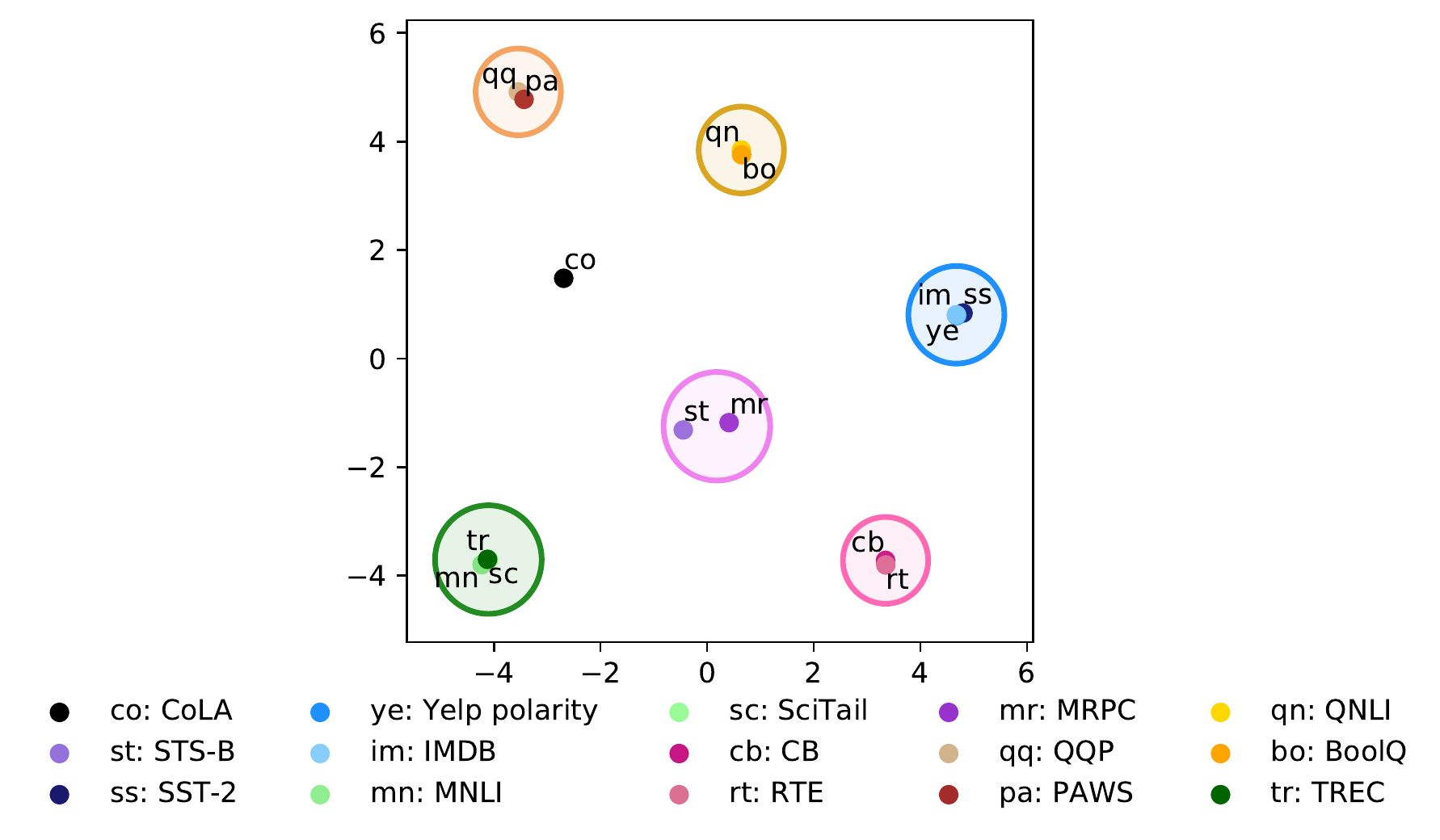}}
\caption{Visualization of learned task embeddings by \methodefficientbase.}
\label{fig:task_embeddings} 
\end{figure}

To analyze what \methodefficientbase has learned about the relations between different tasks, we visualize the learned task embeddings for the models trained with the largest number of samples in Table~\ref{tab:glue_results} and~\ref{tab:transfer_results}. Figure~\ref{fig:task_embeddings} illustrates the 2D vector projections of task embeddings using PCA~\citep{wold1987principal}. Interestingly, the observed groupings correspond to similar tasks. This shows that learned task embeddings by \methodefficientbase are meaningful.
For CB, an NLI dataset despite being initialized from MNLI, after few-shot training the task embedding is closest to RTE, another NLI dataset. This is plausible as premises and hypotheses in both the discourse-based CB and the news and Wikipedia-based RTE are more complex compared to MNLI. The sentence similarity dataset STS-B is grouped close to the MRPC paraphrase dataset. CoLA, which focuses on linguistic acceptability is very different from other tasks and is not grouped with any of the observed task embeddings. In addition, the task embeddings for 1) all the sentiment analysis datasets namely SST-2, Yelp polarity, and IMDB; 2) the two large-scale NLI datasets namely MNLI and SciTail; 3) question answering datasets, i.e. BoolQ and QNLI; and 4) paraphrase datasets namely QQP and PAWS are each grouped together.

\section{Related Work}

\paragraph{Multi-task learning:} Multi-task learning, i.e., learning a unified model to perform well on multiple different tasks, is a challenging problem in NLP.  It requires addressing multiple challenges such as catastrophic forgetting, and handling disproportionate task sizes resulting in a model overfitting in low-resource tasks while underfitting in high-resource ones~\cite{arivazhagan2019massively}.~\citet{liu2019multi} proposed Multi-Task Deep Neural Network (MTDNN) for learning from multiple NLU tasks. Although MTDNN obtains impressive results on GLUE, it applies multi-task learning as a form of pretraining followed by task-specific fine-tuning. Concurrently with us,~\citet{tay2021hypergrid} propose a multi-task learning method by training task-conditioned hyper networks; however, their method is 43x less parameter efficient compared to ours. In another line of research,~\citet{clark2019bam} proposed to learn multi-task models with knowledge distillation. \citet{houlsby2019parameter} trained adapters for each task separately, keeping the model fixed. \citet{stickland2019bert} share the model parameters across tasks and introduce task-specific adapter parameters, which is more parameter-inefficient than our method. 

\paragraph{Hypernetworks and contextual parameter generation:} Our work is closely related to hypernetworks~\citep{Ha2017hypernetworks}. 
In a continual learning setup, where tasks are learned sequentially,~\citet{Oswald2020continual} proposed a task-conditioned hypernetwork to generate all the weights of the target model. Our method is substantially more efficient as we do not generate all the weights of the target model but a very small number of parameters for adapter modules to allow the model to adapt to each individual task efficiently. Similarly,~\citet{jin2020language} generate the full model from task-specific descriptions in different domains whereas we efficiently generate only small adapter modules for each task. 

Prior work also proposed meta-learning or Bayesian approaches to generate softmax layer parameters for new settings \cite{Bansal2020learning,ponti2020parameter}. Meta-learning approaches are notoriously slow to train. In addition, generating softmax parameters requires a substantially higher number of parameters, leaves the method unable to adapt the lower layers of the model, and restricts their application to classification tasks.

In contemporaneous work, \citet{ustun2020udapter} proposed a multilingual dependency parsing method based on adapters and contextual parameter generator networks~\citep{platanios2018contextual} where they generate adapter parameters conditioned on trained input language embeddings. Their study is limited to multilingual dependency parsing, while our work studies multi-task learning and applies to several tasks thanks to the general sequence-to-sequence nature of our model. Moreover, their number of trainable parameters is $2.88 \times$ larger than their base model since they employ a contextual parameter generator in each layer. In contrast, we use a single compact hypernetwork allowing us to efficiently condition on multiple tasks and layers of a transformer model. 

\section{Conclusion}
We propose a parameter-efficient method for multi-task fine-tuning. 
Our approach is to train shared hypernetworks to generate task-specific adapters conditioned on the task, layer id, and adapter position embeddings. The shared hypernetworks capture the knowledge across tasks and enable positive transfer to low-resource and related tasks, while task-specific layers allow the model to adapt to each individual task. Extensive
experiments show that our method obtains strong improvement over multi-task learning on the GLUE benchmark, and substantially improves the in-domain task generalization.

\section*{Acknowledgments}
We are grateful to Dani Yogatama, Neil Houlsby, and Colin Raffel for feedback on a draft of this paper. We would like to also thank Adam Paszke, Jamie Kiros, and George Dahl for useful comments and discussions. 

\bibliography{anthology,ref}
\bibliographystyle{acl_natbib}
\clearpage
\appendix

\section{Experimental Details} \label{sec:appendix}

\paragraph{Computing infrastructure:} We run the experiments in Table~\ref{tab:glue_results} on 4 GPUs, and the rest of the experiments on 1 GPU on a heterogeneous cluster with Tesla V100, Tesla A100, Tesla P4, and GTX1080ti GPUs.

\paragraph{Hyperparameters:} We use a batch size of 64  for \basesmall and 32 for \basebase to fit the GPU memory. We set the dimension of the task feature embedding ($\bm{z_{\tau}}$) to $t'=512$, and the dimension of the task embedding ($\bm{I_{\tau}}$) to $t=64$. For low-resource fine-tuning in \textsection\ref{sec:lowresource}, we use reduction factors of $\{16, 32, 64\}$.

\paragraph{Data pre-processing:} We download all datasets from the HuggingFace Datasets library \cite{2020HuggingFace-datasets}. Following~\citet{raffel2019exploring}, we cast all datasets into a sequence-to-sequence format, and recast STS-B as a 21-class classification task by rounding its target scores to their nearest increment of 0.2.

\paragraph{Performance evaluation:} Table~\ref{tab:mem} and ~\ref{tab:time} present the efficiency evaluation in terms of memory, and time for all the methods measured on the GLUE benchmark. We report the time for 1000 training steps.

Our approach has several attractive properties. Our \methodefficientbase approach offers a much better memory usage with low-overhead, while \methodbase and \basebase cause substantial memory overhead. In dealing with large-scale transformer models like T5, efficient memory usage is of paramount importance. Second, in terms of training time, our method
is much faster than \madapterbase. Relative to \basebase, \methodefficientbase increases the training time by 30.49\%, while \madapterbase causes the substantial training time overhead of 84.93\%.

\begin{table}[H]
\centering 
    \begin{tabular}{lll}
    \toprule 
    \textbf{Model} & \textbf{Memory} & $\bm{\Delta \%}$ \\
    \toprule 
    \basebase     & 7.76 (GB) & -\\
    \madapterbase & 5.95 (GB) & -23.32\% \\
    \methodbase & 7.60 (GB) & -2.06\% \\ 
    \methodefficientbase & 5.81 (GB) & -25.13\\
    \bottomrule
    \end{tabular}\vspace{-0.5em}
     \caption{The required memory for all methods. $\bm{\Delta \%}$ is the relative difference with respect to \basebase.}
       \label{tab:mem} 
      \vspace{-0.5em}
\end{table}
\vspace{-0.5em}
\begin{table}[H]
\centering

    \begin{tabular}{lll}
    \toprule 
    \textbf{Model} &  \textbf{Time} & $\bm{\Delta \%}$ \\
    \toprule 
    \basebase     & 5.51 \hspace{0.5em}(min) & -\\
    \madapterbase &  10.19 (min) & 84.93\%\\
    \methodbase & 7.92 \hspace{0.5em}(min) & 43.74\%\\  
    \methodefficientbase &  7.19 \hspace{0.5em}(min)  & 30.49\%\\
    \bottomrule
    \end{tabular}\vspace{-0.5em}
     \caption{Training time for all methods. $\bm{\Delta \%}$ is the relative difference with respect to \basebase.}
       \label{tab:time} 
      \vspace{-1em}
\end{table}

\paragraph{Impact of adapter's bottleneck size on the performance} Similar to ~\citep{houlsby2019parameter}, adapter's reduction factor needs to be set per dataset. Table~\ref{tab:val_glue_results} shows the validation performance of \methodefficient\! on the GLUE tasks for different adapters' reduction factors.  While the pattern may not be always consistent, generally, smaller datasets seem to benefit more from smaller bottleneck size, i.e., less parameters for adapters, while the opposite is the case for larger datasets, which require more modeling capacity.

\begin{table*}[ht!]
\centering 
\begin{adjustbox}{max width=\textwidth}
\begin{tabular}{l|l|llllllll|l}
\toprule 
\textbf{Model} & \textbf{r}&  \textbf{CoLA} &    \textbf{SST-2} &   \textbf{MRPC} &   \textbf{QQP} &     \textbf{STS-B} & \textbf{MNLI}  &    \textbf{QNLI} & \textbf{RTE} &   \textbf{Avg} \\
\toprule 
\methodefficientsmall & 8&    42.13 &         98.60 &        82.76/87.72 &       90.69/87.55 &            84.92/84.18 &        82.3  &        95.40  &       78.83 &   83.19 \\
\methodefficientsmall & 16&   42.60  &         97.8 &        84.73/89.12 &       88.99/85.33 &            85.69/85.12 &        81.96 &        93.69 &       75.91  &  82.81 \\ 
\methodefficientsmall & 32&   49.90  &         96.00   &        83.74/88.50  &       89.29/85.79 &            85.99/85.41 &        81.28 &        91.79 &       72.99  &  82.79\\ 
\midrule 
\methodefficientbase & 8 &   54.86 &         97.30 &        88.18/91.55 &       94.59/92.91 &            89.77/89.69 &        85.89 &        96.10  &       84.67 & 87.77 \\ 
\methodefficientbase & 16 & 53.83 &         98.00   &        88.18/91.61 &       94.89/93.33 &            90.12/89.65 &        85.94 &        96.50  &       83.94  & 87.82  \\ 
\methodefficientbase & 32&          55.58 &         97.20 &        89.66/92.42 &       93.19/91.08 &            88.96/88.57 &        85.82 &        94.19 &       81.75 & 87.13 \\ 
\bottomrule
\end{tabular}
\end{adjustbox} \vspace{-0.5em}
\caption{Validation performance of \methodefficient\! on the GLUE tasks for different reduction factors $r=\{8, 16, 32\}$. For MNLI, we report accuracy on the matched validation set. For MRPC and QQP, we report accuracy and F1. For STS-B, we report Pearson and Spearman correlation coefficients. For CoLA, we report Matthews correlation. For all other tasks, we report accuracy.  
\label{tab:val_glue_results}}\vspace{-1em}
\end{table*}

\end{document}